\documentclass{article}

\usepackage{arxiv}

\usepackage[utf8]{inputenc} 
\usepackage[T1]{fontenc}    
\usepackage{hyperref}       
\usepackage{url}            
\usepackage{booktabs}       
\usepackage{amsmath,amssymb,amsfonts}       
\usepackage{nicefrac}       
\usepackage{microtype}      
\usepackage{lipsum}
\usepackage{fancyhdr}       
\usepackage{graphicx}       
\usepackage{rotating}
\usepackage{booktabs} 
\usepackage{stfloats}
\usepackage{float}
\usepackage{array}
\usepackage{csvsimple}
\usepackage{siunitx}
\usepackage{diagbox}
\usepackage{ragged2e}
\usepackage{svg}
\usepackage{algorithmic}
\usepackage{algorithm}
\usepackage{graphicx}
\usepackage[normalem]{ulem}
\usepackage{multirow}
\usepackage{multicol}
\usepackage{rotating}
\usepackage{changepage}
\usepackage{etoolbox}
\usepackage{fancyhdr}

\usepackage{tabularx}
\usepackage{adjustbox}
\usepackage{pifont}

\usepackage{textcomp}
\usepackage{xcolor}

\usepackage[
    backend=biber,natbib,
    style=ieee
]{biblatex}

\graphicspath{{media/}}     

\usepackage{subfiles}
\title{Advancing Long-Term Multi-Energy Load Forecasting with Patchformer: A Patch and Transformer-Based Approach
}

\author{
  Qiuyi Hong \\
  School of Mathematics, Statistics and Actuarial Science \\
  University of Essex \\
  United Kingdom\\
  \texttt{qh20439@essex.ac.uk} \\
   \And
  Fanlin Meng \\
  Alliance Manchester Business School \\
  University of Manchester \\
  United Kingdom\\
  \texttt{fanlin.meng@manchester.ac.uk} \\
  \And
  Felipe Maldonado \\
  School of Mathematics, Statistics and Actuarial Science \\
  University of Essex \\
  United Kingdom\\
  \texttt{felipe.maldonado@essex.ac.uk} \\
}

\begin{document}
\maketitle

\begin{abstract}
In the context of increasing demands for long-term multi-energy load forecasting in real-world applications, this paper introduces Patchformer, a novel model that integrates patch embedding with encoder-decoder Transformer-based architectures. To address the limitation in existing Transformer-based models, which struggle with intricate temporal patterns in long-term forecasting, Patchformer employs patch embedding, which predicts multivariate time-series data by separating it into multiple univariate data and segmenting each of them into multiple patches. This method effectively enhances the model's ability to capture local and global semantic dependencies. The numerical analysis shows that the Patchformer obtains overall better prediction accuracy in both multivariate and univariate long-term forecasting on the novel Multi-Energy dataset and other benchmark datasets. In addition, the positive effect of the interdependence among energy-related products on the performance of long-term time-series forecasting across Patchformer and other compared models is discovered, and the superiority of the Patchformer against other models is also demonstrated, which presents a significant advancement in handling the interdependence and complexities of long-term multi-energy forecasting. Lastly, Patchformer is illustrated as the only model that follows the positive correlation between model performance and the length of the past sequence, which states its ability to capture long-range past local semantic information.    
\end{abstract}


\section{Introduction}
\subsection{Background}
The development of Integrated Multi-Energy Systems (IMES) marks a significant evolution in the energy sector, which reflects a shift towards more diversified, efficient, and sustainable energy management practices. IMES encompasses a blend of various energy sources and forms, including electricity, gas, heat, cooling and renewables, and it integrates into a cohesive system. This integration facilitates a more holistic energy production, distribution, and consumption framework. The evolution of IMES reflects a growing recognition of the need for more resilient and adaptable energy systems, especially in the face of escalating global energy demands and the need for improving collective technical, economic, and environmental performance \cite{mancarella2016modelling}. IMES has become an important strategic development direction in the energy field to deal with the global challenges in the fossil energy crisis, climate changes and environmental pollution.
Central to the effective operation of IMES is the role of energy forecasting. Precise prediction plays a crucial role in overseeing these complex systems, guaranteeing that energy generation and supply correspond with demand trends. This alignment is essential for maximising the economic and environmental benefits of IMES as well as for their operational efficiency. Accurate load forecasting contributes to the economic and environmental sustainability of energy systems by optimising resource allocation and lowering operating costs. For instance, \cite{ranaweera1997economic} shows that the annual economic loss could be up to 10 million pounds when every percentage point increases in the error of electricity load forecasting in the United Kingdom. In addition, \cite{wang2020multi} also illustrates that when the forecasting error is decreased by 1\%, the total energy consumption of 58 million MW/h can be saved in one year in China. Moreover, effective energy load forecasting promotes the integration of renewable energy sources, assisting in the reduction of carbon/ greenhouse gas emissions and furthering the aims of sustainable energy from an environmental standpoint. The prediction's reliability and accuracy are critical for the development of future energy systems to be sustainable and able to satisfy energy demands economically.

\subsection{Literature Review}
The methods of time-series forecasting can be categorised into two groups: 1) traditional methods represented by time-series analysis and regression analysis. 2) artificial intelligence methods represented by machine learning and deep learning. Autoregressive integrated moving-average (ARIMA) \cite{box2015time, box1974some} is one of the most popular models in the former group. It predicts the data by obtaining the fitting equations for time series data and other variables. However, the traditional forecasting methods are mainly based on linear analysis and have limited capacity to deal with nonlinear problems \cite{wang2020multi}. The latter group of time-series forecasting methods includes various machine learning and deep learning techniques, such as Recurrent Neural Networks (RNN) \cite{rangapuram2018deep}, Long Short-Term Memory (LSTM) networks \cite{elsworth2020time}, Gated Recurrent Units (GRU) \cite{zhang2017time, jo2021improved}, and Transformer-based methods \cite{kitaev2020reformer, zhou2021informer, wu2021autoformer, zhou2022fedformer, nie2023a}, which have significantly enhanced forecasting capabilities. Among these, Transformer-based models stand out for their ability to handle large datasets and capture complex temporal relationships in multivariate data, offering substantial improvements over other techniques. This is due to their parallelisability and attention mechanism. In particular, the vanilla transformer model \cite{vaswani2017attention} uses a scaled dot product attention mechanism to calculate the point-wise correlation between two different data points. However, applying the full attention mechanism to long-term time-series forecasting (LTTSF) is computationally expensive due to its quadratic complexity in terms of the length of input sequence $L$, which makes it challenging to handle long-term sequences. To overcome this issue, many improved models for LTTSF have been developed. For instance, Reformer \cite{kitaev2020reformer} has been designed to enhance the efficiency for training on long sequences by introducing the Locality-Sensitive Hashing attention to reduce the computational complexity from $\mathcal{O}(L^2)$ to $\mathcal{O}(L\log L)$. Informer model \cite{zhou2021informer}, which also achieves $\mathcal{O}(L\log L)$ complexity, introduces a ProbSparse self-attention mechanism, self-attention distilling, and a generative style decoder. These features collectively enhance the model's efficiency and prediction capacity, making it a robust solution for LTTSF. Although the computational complexity has been reduced, the above models still use the point-wise attention mechanism to understand the correlation between two data points. This may not be the preferred mechanism in time-series forecasting compared to natural language processing (NLP). It is because, unlike a word in a sequence sentence, a single time step in a time series does not have semantic meaning. Therefore, the point-wise correlation cannot capture the input sequence's local semantic information or pattern, which results in poor LTTSF performance. A few models have been developed to address the problem. For instance, Autoformer \cite{wu2021autoformer} incorporates a decomposition architecture with an Auto-Correlation mechanism calculated by Fast Fourier Transforms (FFT) based on series periodicity. It focuses on discovering dependencies and aggregating representations at the sub-series level. This mechanism is more efficient and accurate than traditional self-attention mechanisms, especially for long-term forecasting. Similarly, FEDformer \cite{zhou2022fedformer} also introduces the seasonal-trend decomposition and Frequency Enhanced Attention block with Discrete Fourier Transform (DFT) to capture the sub-series level correlation in time-series sequences. Moreover, inspired by the Vision Transformer (ViT) \cite{dosovitskiy2021an} which truncates each image into $16 \times 16$ patches before feeding it into the vanilla transformer model, and the following influential work BEiT \cite{bao2022beit}, PatchTST \cite{nie2023a} segments time series into subseries-level patches as input tokens, which is designed to retain local semantic information, and employs channel-independence, where each channel contains a single univariate time series sharing the same embedding and Transformer weights which benefits for multivariate time-series forecasting. This design enhances long-term forecasting accuracy significantly compared to state-of-the-art Transformer models, reduces computation and memory usage, and allows the model to attend to a more extended history. This paper proposes a novel Transformer-based model which integrates the patch embedding mechanism and vanilla transformer's encode-decoder architecture to improve the LTTSF performance. 

The methods for load forecasting also consist of both traditional and artificial intelligence methods. For instance, \cite{lee2011short, fang2016evaluation} apply ARIMA and its variant seasonal autoregressive integrated moving-average (SARIMA) for load forecasting problems. \cite{shi2017deep} proposes a novel pooling-based deep RNN for household load forecasting, which batches a group of customers' load profiles in a pool of inputs. \cite{kong2017short} introduces an LSTM RNN-based framework to forecast the highly volatile and uncertain electric load of an individual energy customer. A novel short-term load forecasting method based on attention mechanism, rolling update and bi-directional long short-term memory (Bi-LSTM) neural network is proposed for short-term electricity load forecasting in \cite{wang2019bi}. 
One of the characteristics of multi-energy data is the interdependence among each energy. Forecasting models have been built to capture the interdependence to enhance the forecasting accuracy in the literature. For instance, \cite{wang2020multi} proposes an encoder-decoder model based on LSTM, considering the high-dimensional temporal dynamic characteristic. To capture the cross-coupling characteristic, a coupling feature matrix for multi-energy load is established. \cite{zhang2021short} presents a convolutional neural network (CNN)-Seq2Seq model with an attention mechanism based on a multi-tasking learning method for a short-term multi-energy load forecasting, which considering temperature, humidity, wind speed, and the coupling relationship of multi-energy. An improved multi-energy forecasting method, which uses a CNN-Attention-LSTM model based on federated learning to predict multi-energy load in the integrated energy microgrid is proposed in \cite{zhang2022federated}. In time-series forecasting literature, the prediction length for LTTSF typically ranges from 96 to 720 time steps with hourly data \cite{nie2023a, wu2021autoformer, zhou2022fedformer, zhang2022crossformer}. On the other hand, short-term energy load forecasting literature usually predicts for a few days or weeks \cite{kuo2018high, wang2019bi, zhang2021short, cen2024multi}, while long-term energy load forecasting literature extends to months or years \cite{mohammed2022adaptive, khuntia2018long, lindberg2019long}. However, the daily or monthly data used in energy load prediction may have fewer time steps to predict than LTTSF literature. To predict multi-energy load with hourly data, this paper follows the definition of long-term in LTTSF literature. Furthermore, long-term multi-energy load forecasting has witnessed a paradigm shift in recent years with the development of Transformer-based models. Many Transformer-based models are developed for multi-energy load forecasting in the literature. For instance, A one-encoder multi-decoder multi-task model is developed in \cite{wang2022transformer} to capture the joint relationships among different energies. A similar idea is also adopted in \cite{wang2023probabilistic} with the novel Bayesian multi-head attention mechanism. Despite the growing body of research on Transformer-based models in multi-energy forecasting, a research gap remains in the literature. To the best of our knowledge, there has been no exploration of Transformer-based models that incorporate patch embedding techniques in long-term multi-energy load forecasting, which has shown great promise in other domains, such as NLP and computer vision (CV), for its ability to capture local contextual information and reduce computational complexity. \cite{cen2024multi} proposes a PatchTCN-TST model, which applies a patching approach but only for short-term multi-load energy forecasting. The absence of patch embedding-based Transformer models in long-term multi-energy load forecasting is a significant neglect. Such models have the potential to enhance the model's ability to process and learn from multivariate time series data, capture local and global semantic information, and provide a deep understanding of energy consumption patterns. This approach could lead to more accurate and robust forecasting models, which are essential for effective energy management and planning in the face of increasing demand and the growing complexity of future energy systems.

\subsection{Contributions}
This paper introduces a novel Transformer-based model that integrates patch embedding techniques for long-term multi-energy load forecasting. This model, which we have named the Patchformer, is designed to address the specific challenges of forecasting energy loads over extended periods. By processing the multivariate time series data into multiple univariate data and segmenting individual univariate data into patches, the Patchformer offers a unique approach to understanding and predicting energy consumption patterns. We believe this model represents an advancement in energy forecasting, filling a critical gap in the existing literature and improving the accuracy and efficiency across long-term forecasting models. The key contributions of this paper are outlined as follows:

\begin{itemize}
    \item Innovative Model Architecture: The Patchformer is designed to integrate the Patch Embedding block from PatchTST and the encoder-decoder structure from the vanilla transformer model. This Patch Embedding block treats each channel of the multivariate time series as a distinct univariate input and segments it into subseries-level patches. This approach captures local semantic information within each univariate time series and learns inter-channel relationships more effectively via a channel-independent approach, where each channel shares the same embedding and Transformer weights, enhancing efficiency in multivariate time-series forecasting. In addition, with its multi-head attention mechanisms, the encoder-decoder structure facilitates the importation of comprehensive information from the encoder to the decoder, potentially improving forecasting accuracy.

    \item First of its kind for Long-Term Multi-Energy Load Forecasting: To the best of our knowledge, the Patchformer is the first Transformer-based model employing a patch embedding method for long-term multi-energy load forecasting. This approach effectively addresses the complexities of predicting multi-energy load over extended periods and captures the interdependence among different energies (e.g., electricity, gas and heat) and other energy-related products (e.g., greenhouse gas (GHG)).

    \item Comprehensive Numerical Analysis: Experiments show the Patchformer model achieves better performance against other state-of-the-art Transformer-based models for multivariate long-term forecasting in the Multi-Energy dataset and six other benchmark datasets. In addition, the model also procures higher accuracy in univariate long-term forecasting when predicting the load of electricity and gas. Moreover, the numerical analysis also illustrates the positive effect of the interdependence among energies and energy-related products on the performance of the forecasting in the Multi-Energy dataset across Patchformer and other models via comparing the model accuracy between predicting the electricity, gas load and GHG emissions all at once and the average of the individual predictions. Lastly, the experiment demonstrates the distinct positive correlation between Patchformer’s performance and the past sequence length, which shows its ability to capture long-range past local semantic information.
\end{itemize}

\subsection{Paper Organisation}
The remainder of this paper is organised as follows. Section \ref{sec:Model Architecture} illustrates the proposed Patchformer mode architecture in detail. In Section \ref{sec:Numerical Analysis}, numerical analysis has been developed and evaluated the performance of the Patchformer as well as other state-of-the-art Transformer-based models in different types of datasets, including one novel multi-energy dataset and six public benchmark datasets. In addition, the multi-energy analysis is also illustrated in the section. Lastly, Section \ref{sec:Conclusion} concludes the paper and discusses future work.

\section{Model Architecture} \label{sec:Model Architecture}
\begin{figure}
	\centering
	\includegraphics[width=1\linewidth]{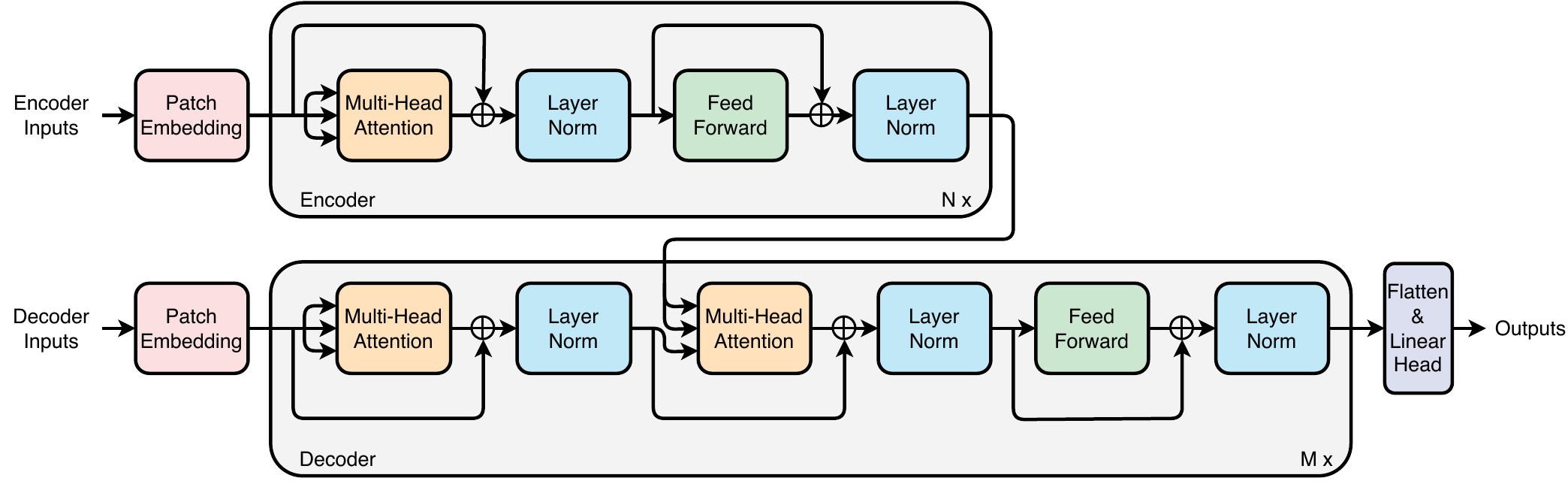}
	\caption{Patchformer Architecture}
	\label{fig:Patchformer}
\end{figure} 

\begin{figure}
	\centering
	\includegraphics[width=1\textwidth]{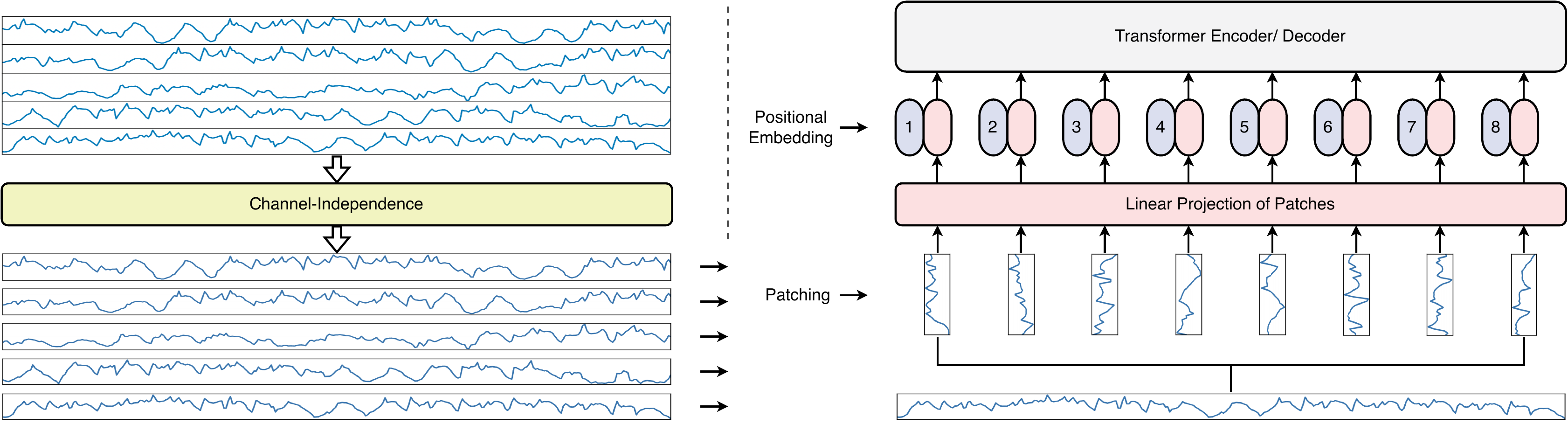}
	\caption{Patch Embedding}
	\label{fig:Patch Embedding}
 \end{figure}

This section introduces the Patchformer model architecture, which is shown in Figure \ref{fig:Patchformer}. Section \ref{sec:Model Overview} gives a high-level overview of the Patchformer model. Section \ref{sec:Patch Embedding} presents one of the core concepts of the model, which is the patching embedding approach. Section \ref{sec:Multi-Head Attention Block},\ref{sec:Layer Normalisation Block} and \ref{sec:Feed Forward Block} depict each of the building blocks for the encoder and decoder. Finally, the structure of the encoder, decoder, and linear head are illustrated in Section \ref{sec:Encoder},\ref{sec:Decoder} and \ref{sec:Flatten and Linear Head}, respectively.    

\subsection{Model Overview} \label{sec:Model Overview}
The model consists of an encoder and a decoder, while the maximum number of layers in the encoder and decoder are $N$ and $M$, respectively. For the multivariate time-series forecasting, the past time sequence is denoted as $\mathcal{X} \in \mathbb{R}^{I \times C}$ with a total of length $I$ and channels $C$. $\mathcal{X}^{c}_{t:I}$ represents a data point at time $t$ and channel $c$. The future/prediction time sequence is represented as $\mathcal{Y} \in \mathbb{R}^{O \times C}$ with the total length $O$. Through the encoder, the information of its inputs is imported into the decoder to provide extra past information for the decoder to predict future sequences.

\subsection{Patch Embedding}\label{sec:Patch Embedding}
The past multivariate time sequence $\mathcal{X}$ is split as $C$ univariate time sequences $\mathcal{X}^{c} \in \mathbb{R}^{1 \times I}$ before patching, which is the representation of the channel-independence. Each $\mathcal{X}^{c}$ is then segmented into multiple patches $\mathcal{X}^{c}_{z:Z} \in \mathbb{R}^{Z \times P}$ by the patch length $P$ and stride $S$ which is similar to the idea in CNN. Therefore, the total number of patches is calculated by $Z = \lfloor \frac{I - P}{S} \rfloor + 2$. Notice that the patching method always pads extra $P$ time steps with the last value of the past time sequence to ensure all time-series data are in patches \cite{nie2023a}. After patching, a one-dimensional univariate time series data $\mathcal{X}^{c}$ is converted to a two-dimensional matrix $\mathcal{P}^{c}_{patch}$ in which each row represents a patch. In addition, value embedding which projects $\mathcal{P}^{c}_{patch}$ from $\mathbb{R}^{Z \times P}$ dimensional space into $\mathbb{R}^{Z \times D}$ dimensional space and positional embedding is applied to optimise the patch representation and ordering. Figure  \ref{fig:Patch Embedding} shows the procedure of the patch embedding approach. Lastly, the patch embedding block can be represented as $\mathcal{P}^{c} = \text{PatchEmbed}(\mathcal{X}^{c})$ which are illustrated as follows: 
\begin{equation}\tag{1}\label{1}
\begin{aligned}
    & \mathcal{P}^{c}_{patch} = \text{Patching}(\text{Padding}(\mathcal{X}^{c})) \\ 
    & \mathcal{P}^{c} = \mathcal{P}^{c}_{patch} W_{valEmbed} + \text{PosEmbed}(\mathcal{P}^{c}_{patch})
\end{aligned}
\end{equation}
\noindent where $W_{valEmbed} \in \mathbb{R}^{P \times D}$ is a learnable weight for value embedding and $\text{PosEmbed}(.)$ denotes positional embedding. $\mathcal{P}^{c} \in \mathbb{R}^{Z \times D}$ is the patch embedded output.

\subsection{Multi-Head Attention Block}\label{sec:Multi-Head Attention Block}

\begin{figure}
	\includegraphics[width=1\linewidth]{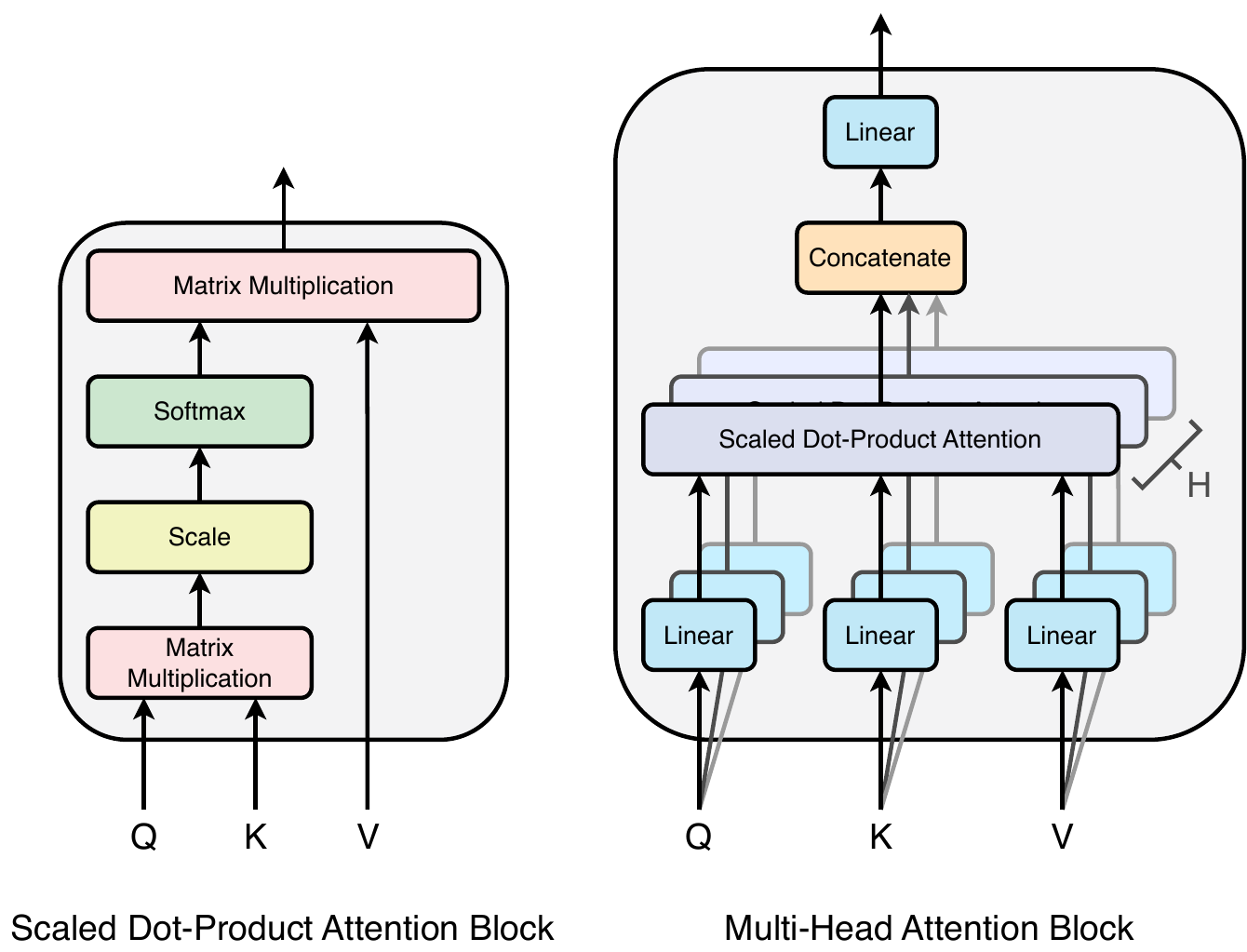}
	\caption{Scaled Dot-Product Attention Block (left) and Multi-Head Attention Block (right), including $H$ attention layers (heads). }
	\label{fig:multi-head attention}
\end{figure}

The Patchformer employs the vanilla transformer's multi-head attention mechanism \cite{vaswani2017attention} to learn the complex local semantic information among patches. Figure \ref{fig:multi-head attention} shows the scaled dot product attention and multi-head attention mechanisms. In particular, the attention layer is used to calculate the attention score by applying the softmax function to the dot product of the similarity between Query $\mathcal{Q}^{c}_{h}$ and Key $\mathcal{K}^{c}_{h}$. The result of the dot product is then scaled by $\sqrt{d_k}$. The attention score is calculated by the product between the scaled dot product and Value $\mathcal{V}^{c}_{h}$. All the $\mathcal{Q}^{c}_{h}$, $\mathcal{K}^{c}_{h}$ and $\mathcal{V}^{c}_{h}$ are computed from the dot product between input $\mathcal{P}^{c}$ and $W^{\mathcal{Q}}$, $W^{\mathcal{K}}$ and $W^{\mathcal{Q}}$, respectively. The multi-head attention block consists of $H$ attention layers, namely, heads, which obtain the multi-head attention score by calculating the dot product between the concatenated attention scores and $W^{O}$. The multi-head attention block $\mathcal{P}^{c}_{attn} = \text{MultiHead}(\mathcal{P}^{c})$ are formulated as follows:
\begin{equation}\tag{2}\label{2}
    \begin{aligned}
        & \mathcal{Q}^{c}_{h} = \mathcal{P}^{c} W^{\mathcal{Q}} \\ 
        & \mathcal{K}^{c}_{h} = \mathcal{P}^{c} W^{\mathcal{K}} \\
        & \mathcal{V}^{c}_{h} = \mathcal{P}^{c} W^{\mathcal{V}} \\
        & \mathcal{H}^{c}_{h} = \text{Attention}(\mathcal{Q}^{c}_{h}, \mathcal{K}^{c}_{h}, \mathcal{V}^{c}_{h}) = \text{Softmax}(\frac{\mathcal{Q}^{c}_{h} \mathcal{K}^{c^{T}}_{h}}{\sqrt{d_k}})\mathcal{V}^{c}_{h} \\ 
        & \mathcal{P}^{c}_{attn} = \text{Concat}(\mathcal{H}^{c}_{1},...,\mathcal{H}^{c}_{H}) W^{O}
    \end{aligned}
\end{equation}
\noindent where the trainable weights $W^{\mathcal{Q}}, W^{\mathcal{K}} \in \mathbb{R}^{D \times d_k}$ and $W^{\mathcal{V}} \in \mathbb{R}^{D \times D}$ result in $\mathcal{Q}^{c}_{h}, \mathcal{K}^{c}_{h} \in \mathbb{R}^{Z \times d_k}$ and $\mathcal{V}^{c}_{h}, \mathcal{H}^{c}_{h} \in \mathbb{R}^{Z \times D}$. Lastly, the multi-head attention output $\mathcal{P}^{c}_{attn} \in \mathbb{R}^{Z \times D}$ is obtained by projecting the row-wise concatenation among all heads $\mathcal{H}^{c}_{h}, \forall h \in \{1,...,H\}$ with weight $W^{o} \in \mathbb{R}^{H*Z \times D}$.

\subsection{Layer Normalisation Block}\label{sec:Layer Normalisation Block}
Layer normalisation is a technique designed to normalise the inputs across the features for each data sample in a mini-batch. Unlike batch normalisation, which normalises across the batch dimension, layer normalisation performs normalisation for each individual sample. For a given sub-layer output $\mathcal{P}^{c}_{attn}$, layer normalisation first computes the mean $\mu^{c}$ and standard deviation $\sigma^{c}$ for each data sample independently. The output of the sub-layer is then normalised by subtracting the mean and dividing by the standard deviation. After normalisation, the process applies two learnable parameters, typically denoted as $\gamma^{c}$ and $\beta^{c}$, which scale and shift the normalised value, respectively. These are trainable parameters which are learned during the training process. The layer normalisation is denoted as $\mathcal{P}^{c}_{norm} = \text{LayerNorm}(\mathcal{P}^{c}_{attn})$ and the detailed process is shown below:
\begin{equation}\tag{3}\label{3}
\begin{aligned} 
    & \mu^{c} = \frac{1}{Z \times D} \sum^{Z}_{z = 1} \sum^{D}_{d = 1} \mathcal{P}^{c}_{attn,z,d} \\ 
    & \sigma^{c} = \sqrt{\frac{1}{Z \times D} \sum^{Z}_{z = 1} \sum^{D}_{d = 1} (\mathcal{P}^{c}_{attn,z,d} - \mu^{c})} \\ 
    & \mathcal{P}^{c}_{norm} = \gamma^{c}(\frac{\mathcal{P}^{c}_{attn} - \mu^{c}}{\sigma^{c}}) + \beta^{c}
\end{aligned}
\end{equation}
Layer normalisation helps stabilise the training process and enables the training of deeper models by mitigating the vanishing or exploding gradient problems. It can lead to faster convergence in training, which is crucial for complex models like transformers that have a large number of parameters.

\subsection{Feed Forward Block}\label{sec:Feed Forward Block}
The Feed Forward block includes two fully connected feed-forward networks (FFNs) with the ReLU activation function in between.
\begin{equation}\tag{4}\label{4}
    \begin{aligned}
        & \mathcal{P}^{c}_{ffn} = \text{Max}(0, \mathcal{P}^{c}_{norm} W^{ffn}_{1} + b_1) W^{ffn}_{2} + b_2
    \end{aligned}
\end{equation}
\noindent where the weights in the FFNs are $W^{ffn}_{1} \in \mathbb{R}^{D \times d_{ff}}$ and $W^{ffn}_{2} \in \mathbb{R}^{d_{ff} \times D}$, respectively. $b_1, b_2 \in \mathbb{R}^{Z}$ denote the bias terms of the FFNs. The feed forward block is summarised by $\mathcal{P}^{c}_{ffn} = \text{FeedForward}(\mathcal{P}^{c}_{norm})$ with the output dimension $(Z, D)$ unchanged compared to its input. 

\subsection{Encoder}\label{sec:Encoder}
With all the above blocks, the Patchformer encoder $\mathcal{P}^{c, l}_{en} = \text{Encoder}(\mathcal{P}^{c, l-1}_{en})$ can be summarised as below: 
\begin{equation}\tag{5}\label{5}
\begin{aligned}
    & \mathcal{P}^{c, l}_{en, 1} = \text{LayerNorm}(\text{MultiHead}(\mathcal{P}^{c, l-1}_{en}) + \mathcal{P}^{c, l-1}_{en}) \\ 
    & \mathcal{P}^{c, l}_{en, 2} = \text{LayerNorm}(\text{FeedForward}(\mathcal{P}^{c, l}_{en, 1}) + \mathcal{P}^{c, l}_{en, 1})
\end{aligned} 
\end{equation}
\noindent where the $\mathcal{P}^{c, l}_{en, 1}$ and $\mathcal{P}^{c, l}_{en, 2}$ represent the outputs of the first and the second layer normalisation blocks, respectively. In addition, the output of the $l$-th encoder layer $\mathcal{P}^{c, l}_{en} = \mathcal{P}^{c, l}_{en, 2}, \forall l \in \{ 1, ..., N \}$ and $\mathcal{P}^{c, 0}_{en} = \mathcal{P}^{c}_{en}$ which is converted from the encoder inputs $\mathcal{X}^{c}_{en}$.

\subsection{Decoder}\label{sec:Decoder}
Patchformer decoder's input $\mathcal{X}^{c}_{de} \in \mathbb{R}^{(\frac{I}{2} + O) \times 1}$ consists of two parts. The first part comes from the second half of the encoder's inputs $\mathcal{X}^{c}_{en}$, denoted as $\mathcal{X}^{c}_{en, \frac{I}{2}:I}$ to provide the most recent past information to the decoder. The second part of $\mathcal{X}^{c}_{de}$ are all zeros. The detailed formulation is shown below: 
\begin{equation}\tag{6}\label{6}
\begin{aligned}
    & \mathcal{X}^{c}_{de} = \text{Concat}(\mathcal{X}^{c}_{en, \frac{I}{2}:I} + \mathcal{X}_{\text{zero}})
\end{aligned}
\end{equation}
\noindent where $\mathcal{X}_{\text{zero}} \in \mathbb{R}^{O \times 1}$ is used as a placeholder to form the decoder input. 

After Patch Embedding, $\mathcal{P}^{c}_{de} = \text{PatchEmbed}(\mathcal{X}^{c}_{de})$, $\mathcal{P}^{c}_{de}$ is obtained as the input for the decoder. Notice that the inner and encoder-decoder multi-head attentions are designed to capture the local semantic information among input patches. By adopting the encoder's output $\mathcal{P}^{c, N}_{en}$, the decoder $\mathcal{P}^{c,l}_{de} = \text{Decoder}(\mathcal{P}^{c, l-1}_{de}, \mathcal{P}^{c, N}_{en})$ can be summarised as follows: 
\begin{equation}\tag{7}\label{7}
\begin{aligned}
    & \mathcal{P}^{c, l}_{de, 1} = \text{LayerNorm}(\text{MultiHead}(\mathcal{P}^{c, l-1}_{de}) + \mathcal{P}^{c, l-1}_{de}) \\ 
    & \mathcal{P}^{c, l}_{de, 2} = \text{LayerNorm}(\text{MultiHead}(\mathcal{P}^{c, l}_{de, 1}, \mathcal{P}^{c, N}_{en}) + \mathcal{P}^{c, l}_{de, 1}) \\ 
    & \mathcal{P}^{c, l}_{de, 3} = \text{LayerNorm}(\text{FeedForward}(\mathcal{P}^{c, l}_{de, 2}) + \mathcal{P}^{c, l}_{de, 2})
\end{aligned}
\end{equation}
\noindent where $\mathcal{P}^{c, l}_{de} = \mathcal{P}^{c,l}_{de,3}, \forall l \in \{ 1,...,M \}$ represents the outputs for the $l$-th decoder layer. In addition, $\mathcal{P}^{c,0}_{de} = \mathcal{P}^{c}_{de}$. $\mathcal{P}^{c, l}_{de, i}, \forall i \in \{ 1, 2, 3 \}$ denote the outputs of layer normalisation blocks in the $l$-th decoder layer, respectively.  
\subsection{Flatten and Linear Head}\label{sec:Flatten and Linear Head}
The Flatten and Linear Head block is designed first to flatten the output of the decoder $\mathcal{P}^{c,M}_{de}$ from dimension $(Z_{de}, P)$ to $(1, Z_{de}*P)$. Second, the final prediction sequence is obtained by converting the output dimension again to $(1, O)$. The detailed formulations are shown below: 
\begin{equation}\tag{8}\label{8}
\begin{aligned}
    & \mathcal{Y}^{c}_{flattened} = \text{Flatten}(\mathcal{P}^{c,M}_{de}) \\ 
    & \mathcal{Y}^{c} = \mathcal{Y}^{c}_{flattened} W_{y} 
\end{aligned}
\end{equation}
\noindent where $\mathcal{Y}^{c}_{flattened} \in \mathbb{R}^{1\times Z_{de}*P}$, $\mathcal{P}^{c,M}_{de} \in \mathbb{R}^{Z_{de} \times P}$. The final prediction time sequence $\mathcal{Y}^{c} \in \mathbb{R}^{1 \times O}, \forall c \in \{1,...,C\}$ is obtained by a linear transformation with weight $W_{y} \in \mathbb{R}^{Z_{de}*P \times O}$.

\section{Numerical Analysis} \label{sec:Numerical Analysis}
In this section, the performance comparison and evaluation between the proposed Patchformer and other state-of-the-art models: Autoformer \cite{wu2021autoformer}, Crossformer \cite{zhang2022crossformer}, and Transformer \cite{vaswani2017attention} and multi-energy analysis are discussed in detail. Firstly, the datasets which contain a novel Multi-Energy dataset, six public benchmark datasets and experimental setup are introduced in Section \ref{sec:Datasets and settings}. The performance of multivariate forecasting for Patchformer and other models across different datasets are discussed in Section \ref{sec:multivariate forecasting}. Sections \ref{sec:univariate forecasting}-\ref{sec:diff past sequence} analyse the Patchformer performance on the Multi-Energy dataset from various perspectives in detail. In particular, Section \ref{sec:univariate forecasting} studies the univariate forecasting performance among Patchformer and other models on the Multi-Energy dataset by predicting electricity, gas load and GHG emission. In addition, the effect of the interdependence among electricity, gas load and GHG emission on the performance of the LTTSF on the Multi-Energy dataset is illustrated in Section \ref{sec:multi-energy forecasting}. Lastly, in Section \ref{sec:diff past sequence}, the forecasting performance is compared between Patchformer and other models with different past sequence lengths. 

\subsection{Datasets and Experimental Setup}\label{sec:Datasets and settings}
The proposed Patchformer model is evaluated on seven datasets, including the novel and comprehensive Multi-Energy dataset \cite{multi-energy} and other six datasets that are well known and have been utilised as benchmarks, publicly available on \cite{wu2021autoformer}. Here is the description of the seven datasets: $(1)$ \textit{Multi-Energy} dataset records energy-related data collected on the Temple campus at Arizona State University, which includes hourly electricity, gas, and heat load demand, renewable energy generation, and GHG emissions for each building from 24 July 2015 to 12 September 2020. $(2)$ \textit{Exchange} dataset collects the daily exchange rate of eight different countries from 1 January 1990 to 10 October 2010. $(3)$ \textit{Weather} dataset is a collection of 21 meteorological indicators (e.g., air temperature, humidity and precipitation) every 10 minutes in the entire year of 2020. $(4)$ and $(5)$ \textit{ETTh1} and \textit{ETTh2} datasets record the hourly data (e.g., load and oil temperature) from two different electricity transformers from 1 July 2016 to 26 June 2018. Similarly, $(6)$ and $(7)$ \textit{ETTm1} and \textit{ETTm2} datasets collect every 15 minutes data from the two electricity transformers in the same time period. The statistics of the seven datasets are shown in Table \ref{tab:datasets}, which shows the total number of input features and length of the time-series observations. The Patchformer and other models are written in PyTorch and run on Ubuntu 22.04.3 LTS x86\_64 with Intel Xeon (8) @ 2.000GHz and 52GB of RAM. The GPU uses NVIDIA Tesla V100 SXM2 16GB. 

Moreover, in this section, the mean squared error (MSE) and mean absolute error (MAE) are applied as experimental evaluation indicators to reflect the forecasting accuracy of the proposed Patchformer and other comparison models. The definitions of the two evaluation indexes are formulated as follows: 

\begin{equation}\tag{9}\label{9}
\begin{aligned}
    & \text{MSE} = \frac{1}{n}\sum^{n}_{i=1}(y_{i} - \hat{y_{i}})^{2} \\
    & \text{MAE} = \frac{1}{n}\sum^{n}_{i=1}|y_{i} - \hat{y_{i}}|
\end{aligned}
\end{equation}

where $n$ is the total number of the data. The actual and predicted values are denoted as $y_{i}$ and $\hat{y_{i}}$ at the $i$-th time step of the dataset, respectively.

\linespread{1.2}
\begin{table}[t]
\small
\caption{Statistics of datasets.}
\centering
\begin{adjustbox}{width=\textwidth}
\begin{tabular}{  c | c  c  c  c  c c c} 
  \hline
   \text{Datasets} & \text{Multi-Energy} & \text{Exchange} & \text{Weather} & \text{ETTh1} & \text{ETTh2} & \text{ETTm1} & \text{ETTm2}\\ 
   \hline
   \text{Features} & 19 & 8 & 21 & 7 & 7 & 7 & 7\\ 
   \text{Length} & 49415 & 7588 & 52696 & 17420 & 17420 & 69680 & 69680\\
  \hline
\end{tabular}
\end{adjustbox}
\label{tab:datasets}
\end{table}

\begin{figure}[t]
	\centering
	\includegraphics[width=0.9\linewidth]{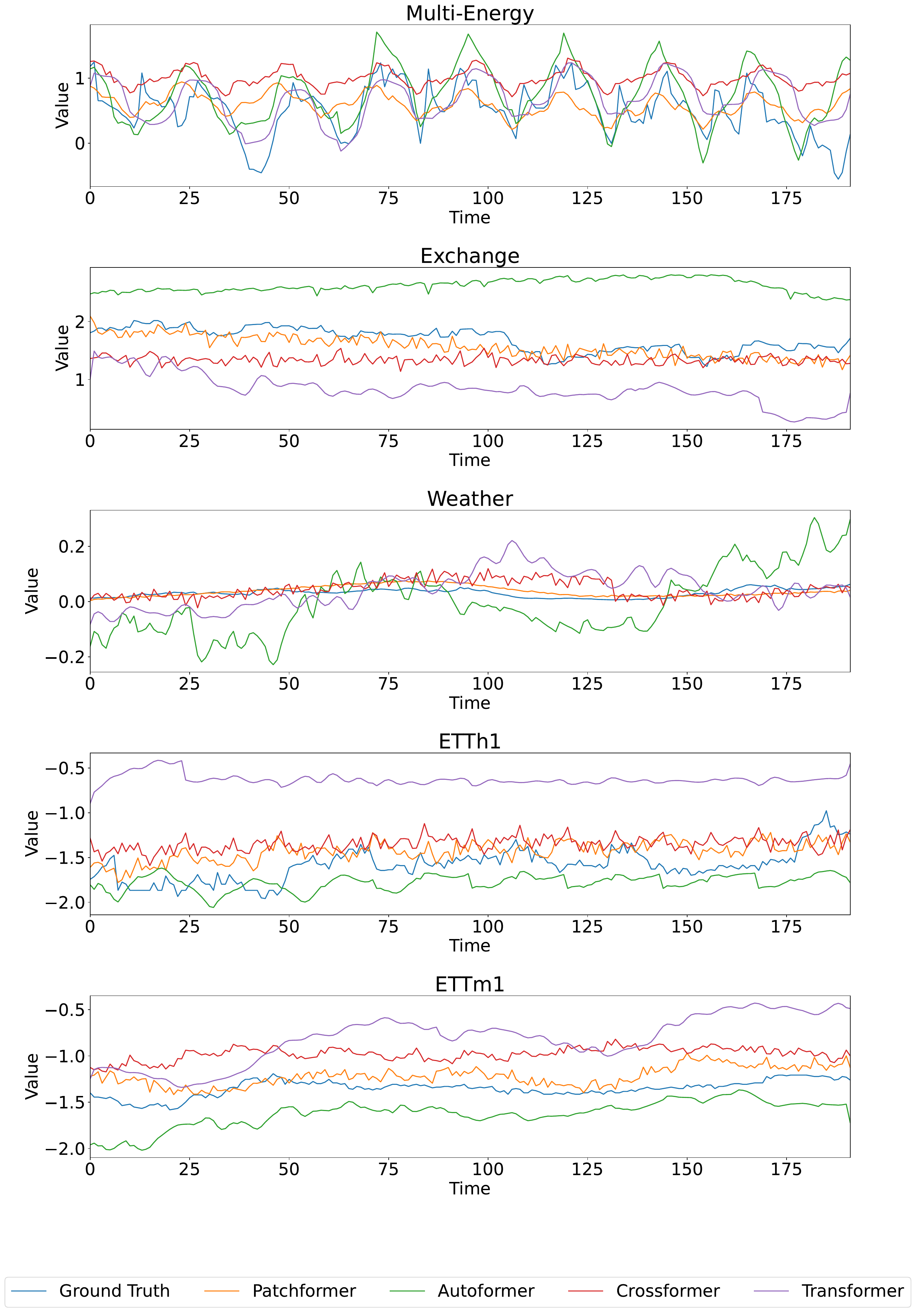}
	\caption{Visualisation of 192 step forecasting on Multi-Energy, Exchange, Weather, ETTh1 and ETTm1 datasets with the past time sequence = 96.}
	\label{fig:time_series}
\end{figure}

\begin{table}[t]
\small
\caption{Hyperparameters of different models for multivariate forecasting.}
\centering
\begin{adjustbox}{width=\textwidth}
\begin{tabular}{  m{2cm} | m{10cm} } 
  \hline
  Models & Hyperparameters  \\ 
  \hline
  Patchformer & patch length = 16, stride = 8, encoder number = 2, decoder number = 1, model dimension = 512, label length = sequence length/2, batch size = 32, head = 8, learning rate = 0.0001, dropout rate = 0.1, Optimiser = Adam, loss function = MSE, epoch = 10  \\ 
  \hline
  Autoformer & topK = 5, encoder number = 2, decoder number = 1, model dimension = 512, label length = sequence length/2, batch size = 32, learning rate = 0.0001, dropout rate = 0.1, Optimiser = Adam, loss function = MSE, epoch = 10 \\ 
  \hline
  Crossformer & topK = 5, encoder number = 2, decoder number = 1, model dimension = 512, label length = sequence length/2, batch size = 32,
learning rate = 0.0001, dropout rate = 0.1, Optimiser = Adam, loss function = MSE, epoch = 10 \\ 
  \hline
  Transformer & encoder number = 2, decoder number = 1, model dimension = 512, label length = sequence length/2, batch size = 32, head = 8, learning rate = 0.0001, dropout rate = 0.1, Optimiser = Adam, loss function = MSE, epoch = 10 \\
  \hline
\end{tabular}
\end{adjustbox}
\label{tab:hyperparameters}
\end{table}

\linespread{1.2}
\begin{table}[t]
\small
\caption{Multivariate time-series forecasting results with Patchformer. We use prediction lengths $\mathcal{Y} \in \{96, 192, 336, 720\}$ and past sequence length $\mathcal{X} = 96$.  The best results are in \textbf{bold}, and the second best is in \uline{underlined}.}
	\centering
	\resizebox{\linewidth}{!}{
		\begin{tabular}{cc|c|cc|cc|cc|ccc}
			\cline{2-11}
			&\multicolumn{2}{c|}{Model}& \multicolumn{2}{c|}{Patchformer}& \multicolumn{2}{c|}{Autoformer}& \multicolumn{2}{c|}{Crossformer}& \multicolumn{2}{c}{Transformer}&  \\
			\cline{2-11}
			&\multicolumn{2}{c|}{Metric}&MSE&MAE&MSE&MAE&MSE&MAE&MSE&MAE\\
			\cline{2-11}
			&\multirow{4}*{\rotatebox{90}{Multi-Energy}}& 96    & \uline{0.062} & \textbf{0.135} & \textbf{0.061} & \uline{0.150} & 0.067 & 0.150 & 0.063 & 0.147  \\
            &\multicolumn{1}{c|}{}& 192   & \textbf{0.073} & \textbf{0.149} & \uline{0.075} & \uline{0.161} & 0.086 & 0.172 & 0.079 & 0.161 \\
            &\multicolumn{1}{c|}{}& 336   & \uline{0.090} & \textbf{0.167} & \textbf{0.083} & \uline{0.173} & 0.165 & 0.243 & 0.153 & 0.217  \\
            &\multicolumn{1}{c|}{}& 720   & \textbf{0.121} & \textbf{0.193} & \uline{0.140} & \uline{0.215} & 0.221 & 0.274 & 0.257 & 0.273  \\
			\cline{2-11}
			&\multirow{4}*{\rotatebox{90}{Exchange}}& 96    & \textbf{0.089} & \textbf{0.217} & \uline{0.153} & \uline{0.285} & 0.256 & 0.367 & 0.550 & 0.579  \\
            &\multicolumn{1}{c|}{} & 192   & \textbf{0.190} & \textbf{0.321} & \uline{0.277} & \uline{0.383} & 0.468 & 0.508 & 0.934 & 0.734  \\
            &\multicolumn{1}{c|}{}& 336   & \textbf{0.387} & \textbf{0.471} & \uline{0.471} & \uline{0.513} & 0.975 & 0.763 & 1.328 & 0.904  \\
            &\multicolumn{1}{c|}{}& 720   & \textbf{1.071} & \textbf{0.769} & \uline{1.107} & \uline{0.818} & 1.620 & 1.029 & 2.565 & 1.336  \\
            \cline{2-11}
			&\multirow{4}*{\rotatebox{90}{Weather}}& 96    & \textbf{0.175} & \textbf{0.231} & 0.342 & 0.385 & \uline{0.177} & \uline{0.242} & 0.353 & 0.412  \\
			&\multicolumn{1}{c|}{}& 192   & \textbf{0.213} & \textbf{0.274} & 0.321 & 0.374 & \uline{0.222} & \uline{0.289} & 0.574 & 0.542  \\
			&\multicolumn{1}{c|}{}& 336   & \textbf{0.263} & \textbf{0.311} & 0.347 & 0.384 & \uline{0.276} & \uline{0.338} & 0.631 & 0.584  \\
			&\multicolumn{1}{c|}{}& 720   & \textbf{0.339} & \textbf{0.369} & 0.415 & 0.418 & \uline{0.372} & \uline{0.411} & 0.850 & 0.686  \\
			\cline{2-11}
			&\multirow{4}*{\rotatebox{90}{ETTh1}}& 96    & \uline{0.425} & \uline{0.444} & 0.529 & 0.487 & \textbf{0.419} & \textbf{0.439} & 0.773 & 0.684  \\
            &\multicolumn{1}{c|}{}& 192   & \textbf{0.484} & \textbf{0.477} & \uline{0.509} & \uline{0.486} & 0.539 & 0.517 & 0.886 & 0.744  \\
            &\multicolumn{1}{c|}{}& 336   & \uline{0.549} & \uline{0.512} & \textbf{0.508} & \textbf{0.494} & 0.709 & 0.638 & 0.966 & 0.770  \\
            &\multicolumn{1}{c|}{}& 720   & \uline{0.603} & \uline{0.566} & \textbf{0.542} & \textbf{0.520} & 0.721 & 0.622 & 1.016 & 0.800  \\
			\cline{2-11}
           &\multirow{4}*{\rotatebox{90}{ETTh2}}& 96    & \textbf{0.342} & \textbf{0.387} & \uline{0.375} & \uline{0.410} & 0.790 & 0.612 & 2.633 & 1.291  \\
            &\multicolumn{1}{c|}{}& 192   & \uline{0.473} & \uline{0.459} & \textbf{0.443} & \textbf{0.449} & 1.830 & 1.041 & 5.961 & 2.007  \\
            &\multicolumn{1}{c|}{}& 336   & \textbf{0.475} & \textbf{0.478} & \uline{0.501} & \uline{0.496} & 1.863 & 1.088 & 5.811 & 1.948  \\
            &\multicolumn{1}{c|}{}& 720   & \uline{0.600} & \uline{0.538} & \textbf{0.496} & \textbf{0.499} & 2.833 & 1.447 & 2.964 & 1.399  \\
			\cline{2-11}
            &\multirow{4}*{\rotatebox{90}{ETTm1}}& 96    & \uline{0.364} & \textbf{0.393} & 0.512 & 0.485 & \textbf{0.362} & \uline{0.403} & 0.725 & 0.620  \\
            &\multicolumn{1}{c|}{}& 192   & \uline{0.411} & \textbf{0.421} & 0.539 & 0.494 & \textbf{0.388} & \uline{0.422} & 0.870 & 0.703  \\
            &\multicolumn{1}{c|}{}& 336   & \textbf{0.437} & \textbf{0.447} & \uline{0.587} & \uline{0.523} & 0.617 & 0.579 & 1.062 & 0.790  \\
            &\multicolumn{1}{c|}{}& 720   & \textbf{0.499} & \textbf{0.482} & \uline{0.650} & \uline{0.535} & 0.931 & 0.722 & 1.063 & 0.789  \\
			\cline{2-11}
            &\multirow{4}*{\rotatebox{90}{ETTm2}}& 96    & \textbf{0.214} & \textbf{0.308} & \uline{0.230} & \uline{0.314} & 0.250 & 0.333 & 0.469 & 0.500  \\
            &\multicolumn{1}{c|}{}& 192   & \uline{0.321} & \uline{0.380} & \textbf{0.281} & \textbf{0.340} & 0.888 & 0.694 & 1.438 & 0.891  \\
            &\multicolumn{1}{c|}{}& 336   & \uline{0.373} & \uline{0.415} & \textbf{0.338} & \textbf{0.373} & 1.451 & 0.850 & 1.154 & 0.818  \\
            &\multicolumn{1}{c|}{}& 720   & \uline{0.592} & \uline{0.529} & \textbf{0.459} & \textbf{0.441} & 2.678 & 1.148 & 2.675 & 1.208  \\
			\cline{2-11}
		\end{tabular}
  }
	\label{tab:comparison}
\end{table}
\linespread{1}

\subsection{Multivariate Forecasting on Different Datasets}\label{sec:multivariate forecasting}
In this section, the performance of multivariate forecasting among different models on the above-mentioned seven datasets is compared. Multivariate forecasting considers the historical data of several variables to forecast one or more of these variables while taking into account the interdependence between multiple input variables. The same number of input and output variables is applied to analyse the performance of the multivariate forecasting in this section. The hyperparameters used in the section are shown in Table \ref{tab:hyperparameters}. The prediction and past sequence length are set to be $\mathcal{Y} \in \{96, 192, 336, 720\}$ and $\mathcal{X} = 96$, respectively. The evaluation results are shown in Table \ref{tab:comparison}. Patchformer consistently outperformed competing models at prediction sequence lengths of  192, 336 and 720 on the Multi-Energy dataset introduced in this paper. Notably, at the 720-step forecast, the Patchformer achieved an MSE of 0.121 and an MAE of 0.193, which is 15.70\% and 11.40\% less than the second-best scores, respectively. It indicates its capability to capture the complex interdependencies among the multiple energy vectors over long-term horizons. In addition, at the extended horizon of 96, 192 and 336 steps, the Patchformer's performance remains competitive, with all the best MSE and MAE, except for two second-best MSE at 96 and 336 steps, demonstrating the model's robustness in long-term forecasting. Its superiority in the multi-energy forecasting domain is critical for modern IMES systems. 

For benchmark datasets in multivariate time series forecasting, the Patchformer exhibited varying degrees of efficacy. In the Exchange Rate dataset, the Patchformer performs the best across all prediction lengths and is optimal at shorter horizons but showed a significant decline as the prediction length increased, with the MSE rising sharply to 1.071 at the 720-step horizon. This suggests a potential vulnerability in the Patchformer's architecture when dealing with the non-stationary and volatile nature of financial time series over long-term periods.

In Weather forecasting, the Patchformer maintained competitiveness and the best performance across all horizons, with the best MSE and MAE for all prediction steps. This indicates the Patchformer's adeptness at modelling environmental time series data, which often have clearer temporal patterns and seasonality.

With the ETTh1, ETTh2, ETTm1, and ETTm2 datasets, which contain the data from electricity transformers, including load and oil temperature, the Patchformer displays either the best or second-best MSE and MAE for all prediction lengths, which indicates the model's robustness and reliability. In particular, Patchformer obtains the best MSE and MAE for all different prediction lengths on the ETTm1 dataset, except for MSE at 96 and 192 steps, which are relatively close to the best results obtained by Crossformer.

Figure \ref{fig:time_series} visualises the time-series predictions of the four models on Multi-Energy, Exchange, Weather, ETTh1 and ETTm1 datasets. In particular, the predicted values of the electricity load on the Multi-Energy dataset are visualised to show the models' performance. As a result, the Patchformer exhibits strong performance across multiple datasets since it efficiently captures the trend and is closest to the ground truth. The experimental results indicate that the Patchformer can handle long-term predictions efficiently, especially in domains where the data has explicit seasonality patterns. With high volatility and irregular patterns, such as financial markets, the Patchformer's long-term forecasting performance can be improved in future work.

\begin{figure}[t]
	\includegraphics[width=1\linewidth]{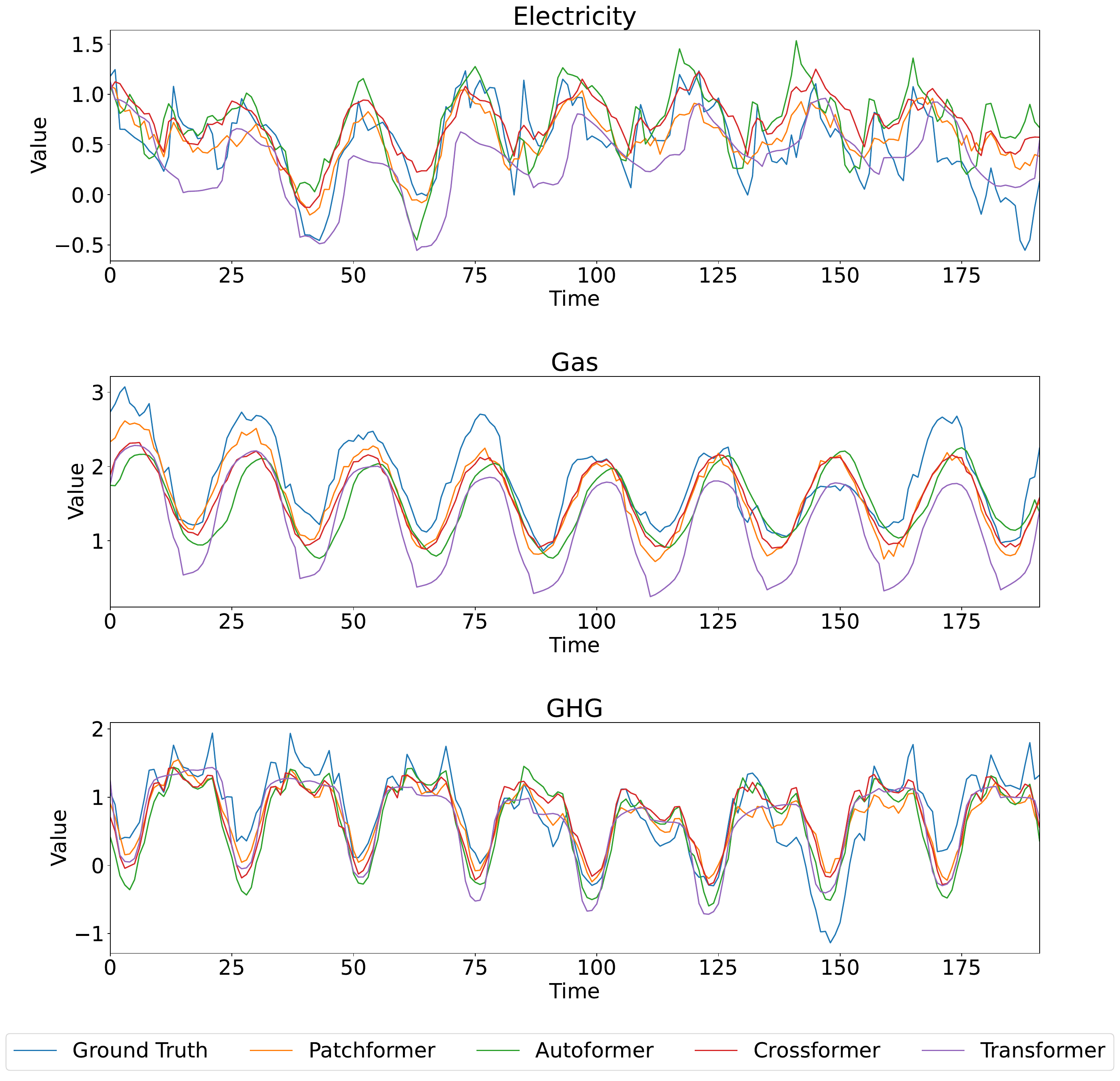}
	\caption{Visualisation of 192 step forecasting on Electricity, Gas and GHG in Multi-Energy dataset with the past sequence length = 336.}
	\label{fig:univariate_forecasting_192}
\end{figure} 

\subsection{Univariate Forecasting} \label{sec:univariate forecasting}

\begin{figure}[t]
	\includegraphics[width=1\linewidth]{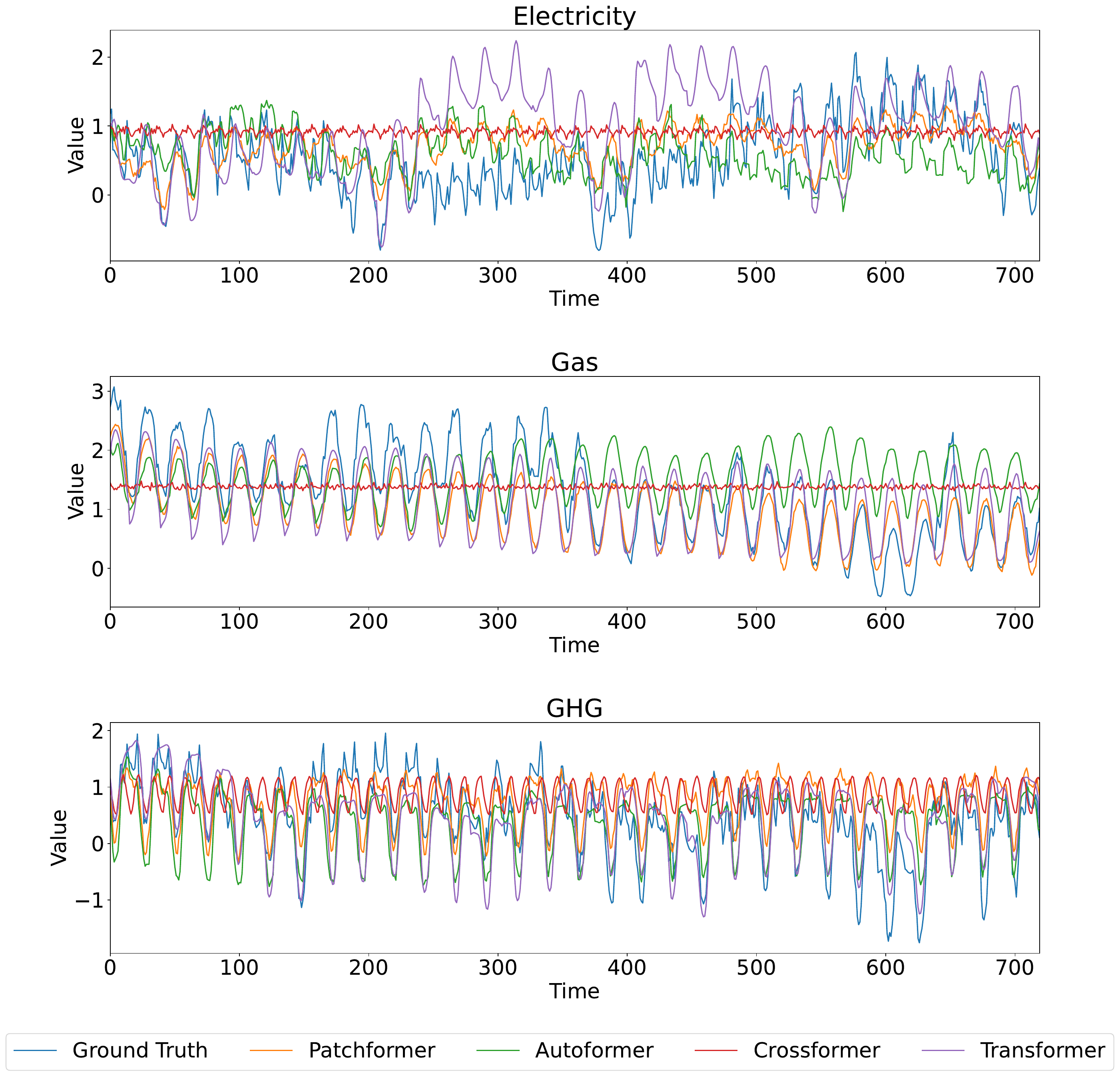}
	\caption{Visualisation of 720 step forecasting on Electricity, Gas and GHG in Multi-Energy dataset with the past sequence length = 336.}
	\label{fig:univariate_forecasting_720}
\end{figure}

In addition to multivariate forecasting, this section presents univariate forecasting results with Patchformer and other comparing models on the Multi-Energy dataset. Univariate forecasting focuses on forecasting the future values of that single variable based on its own historical data. The output is the prediction of future values of the same single variable without considering any other input variables. The features chosen to be analysed are electricity and gas load demand, as well as GHG emissions. The hyperparameters of Patchformer among different prediction lengths are shown in Table \ref{tab:hyperparameters for univariate forecasting}. The past sequence length for all models is fixed at 336 time steps, and the forecasting horizons are 96, 192, 336, and 720 time steps. Table \ref{tab:univariate forecasting} shows univariate forecasting results. Patchformer outperforms all other models on electricity and gas forecasting as it receives the best MSE and MAE results across all prediction lengths. For the forecasting results on the GHG dataset, Patchformer has room for improvement, which implies there may not be a single model that universally outperforms others across all metrics, electricity and gas demand, GHG emissions, and prediction lengths. In addition, Figures \ref{fig:univariate_forecasting_192} and \ref{fig:univariate_forecasting_720} visualise 336 steps past sequence length forecasting with all models on Electricity, Gas and GHG in the Multi-Energy dataset when the prediction lengths are 192 and 720 steps, respectively. The time-series patterns among Electricity, Gas and GHG are shown to be different in both figures. Also, it is fairly obvious to observe that the predictions of Patchformer are the closest to the ground truth value when predicting electricity and gas in Figures \ref{fig:univariate_forecasting_192} and \ref{fig:univariate_forecasting_720}, which indicates the excellent performance of Patchformer. Furthermore, from Figure \ref{fig:univariate_forecasting_720}, the prediction of Crossformer is stated as relatively consistent across time while other models fluctuate dramatically.

\begin{table}[t]
\small
\caption{Hyperparameters of Patchformer for univariate forecasting.}
\centering
\begin{adjustbox}{width=\textwidth}
\begin{tabular}{  m{3cm} | m{10cm} } 
  \hline
  Prediction Length & Hyperparameters  \\ 
  \hline
  96 & patch length = 16, stride = 8, encoder number = 2, decoder number = 1, model dimension = 512, label length = sequence length/2, batch size = 32, head = 16, fully connected layer dimension = 2048, learning rate = 0.0001, dropout rate = 0.1, Optimiser = Adam, loss function = MSE, epoch = 10  \\ 
  \hline
  192 & patch length = 16, stride = 8, encoder number = 2, decoder number = 1, model dimension = 256, label length = sequence length/2, batch size = 32, head = 16, fully connected layer dimension = 1024, learning rate = 0.0001, dropout rate = 0.1, Optimiser = Adam, loss function = MSE, epoch = 10 \\ 
  \hline
  336 & patch length = 16, stride = 8, encoder number = 2, decoder number = 1, model dimension = 512, label length = sequence length/2, batch size = 32, head = 16, fully connected layer dimension = 2048, learning rate = 0.0001, dropout rate = 0.1, Optimiser = Adam, loss function = MSE, epoch = 10 \\ 
  \hline
  720 & patch length = 16, stride = 8, encoder number = 2, decoder number = 1, model dimension = 128, label length = sequence length/2, batch size = 32, head = 16, fully connected layer dimension = 1024, learning rate = 0.0001, dropout rate = 0.1, Optimiser = Adam, loss function = MSE, epoch = 10  \\
  \hline
\end{tabular}
\end{adjustbox}
\label{tab:hyperparameters for univariate forecasting}
\end{table}

\linespread{1.2}
\begin{table}[t]
\small
\caption{Univariate multi-energy forecasting results with Patchformer. We use prediction lengths $\mathcal{Y} \in \{96, 192, 336, 720\}$ and past sequence length $\mathcal{X} = 336$.  The best results are in \textbf{bold} and the second best are \uline{underlined}.}
	\centering
	\resizebox{\linewidth}{!}{
		\begin{tabular}{cc|c|cc|cc|cc|ccc}
			\cline{2-11}
			&\multicolumn{2}{c|}{Model}& \multicolumn{2}{c|}{Patchformer}& \multicolumn{2}{c|}{Autoformer}& \multicolumn{2}{c|}{Crossformer}& \multicolumn{2}{c}{Transformer}&  \\
			\cline{2-11}
			&\multicolumn{2}{c|}{Metric}&MSE&MAE&MSE&MAE&MSE&MAE&MSE&MAE\\
			\cline{2-11}
			&\multirow{4}*{\rotatebox{90}{Electricity}}& 96    & \textbf{0.114} & \textbf{0.262} & 0.176 & 0.326 & \uline{0.116} & \uline{0.264} & 0.154 & 0.304  \\
            &\multicolumn{1}{c|}{}& 192   & \textbf{0.148} & \textbf{0.300} & 0.230 & 0.372 & \uline{0.151} & \uline{0.306} & 0.186 & 0.336 \\
            &\multicolumn{1}{c|}{}& 336   & \textbf{0.181} & \textbf{0.332} & 0.291 & 0.427 & 0.227 & 0.375 & \uline{0.226} & \uline{0.369}  \\
            &\multicolumn{1}{c|}{}& 720   & \textbf{0.272} & \textbf{0.412} & \uline{0.290} & \uline{0.428} & 0.340 & 0.472 & 0.497 & 0.556  \\
			\cline{2-11}
			&\multirow{4}*{\rotatebox{90}{Gas}}& 96    & \textbf{0.071} & \textbf{0.195} & 0.114 & 0.262 & \uline{0.074} & \uline{0.196} & 0.101 & 0.232  \\
            &\multicolumn{1}{c|}{} & 192   & \textbf{0.084} & \textbf{0.221} & 0.121 & 0.269 & \uline{0.095} & \uline{0.230} & 0.120 & 0.262  \\
            &\multicolumn{1}{c|}{}& 336   & \textbf{0.106} & \textbf{0.244} & 0.193 & 0.340 & 0.130 & 0.273 & \uline{0.118} & \uline{0.264}  \\
            &\multicolumn{1}{c|}{}& 720   & \textbf{0.143} & \textbf{0.293} & 0.335 & 0.439 & 0.270 & 0.405 & \uline{0.169} & \uline{0.307}  \\
            \cline{2-11}
			&\multirow{4}*{\rotatebox{90}{GHG}}& 96    & \uline{0.141} & \uline{0.290} & 0.242 & 0.368 & \textbf{0.135} & \textbf{0.285} & 0.172 & 0.323  \\
			&\multicolumn{1}{c|}{}& 192   & 0.202 & 0.354 & 0.217 & 0.369 & \textbf{0.159} & \textbf{0.312} & \uline{0.184} & \uline{0.335}  \\
			&\multicolumn{1}{c|}{}& 336   & 0.280 & 0.413 & \uline{0.226} & \textbf{0.373} & \textbf{0.225} & \uline{0.377} & 0.266 & 0.399  \\
			&\multicolumn{1}{c|}{}& 720   & 0.426 & 0.518 & \textbf{0.311} & \textbf{0.442} & 0.550 & 0.583 & \uline{0.317} & \uline{0.449}  \\
			\cline{2-11}
		\end{tabular}
  }
	\label{tab:univariate forecasting}
\end{table}
\linespread{1}

\subsection{Multi-Energy Forecasting Comparison} \label{sec:multi-energy forecasting}
In this section, the effect of the interdependence among energy-related products on the performance of the time-series forecasting in the Multi-Energy dataset is discussed. The features in the dataset chosen to be analysed are electricity and gas load demand, as well as GHG emissions, since they are highly interrelated in nature. The prediction, past sequence length and hyperparameters of the Patchformer model are identical to Section \ref{sec:univariate forecasting}. The forecasting results are shown in Table \ref{tab:multi-energy forecasting comparison}, in which All-at-Once means to predict electricity and gas load and GHG emission simultaneously, whereas Electricity, Gas and GHG are predicted individually. The feature Average is calculated by the mean of the features Electricity, Gas and GHG. In Table \ref{tab:multi-energy forecasting comparison}, Patchformer outperforms all other three models across all features selected from the Multi-Energy dataset as it achieves the most best MSE and MAE results than other models. In addition, the Patchformer forecasting results at 336 and 720 prediction lengths when predicting electricity, gas load, and GHG emission all at once are better than predicted individually. Furthermore, the difference of Patchformer MSE and MAE between All-at-Once and Average at 96 and 192 prediction lengths are insignificant (MSE: 2.65\% and 1.36\%, MAE: 1.96\% and 0.34\%). Moreover, the results of All-at-Once are generally better than Average for all other models, especially Autoformer and Transformer. Overall, the pattern in which predicting multi-energy results all at once is better than predicting them individually demonstrates the interdependence among electricity, gas load and GHG emission can be captured by Patchformer and other models and improve the forecasting performance.

\linespread{1.2}
\begin{table}[t]
\small
\caption{Multi-energy forecasting results with Patchformer. We use prediction lengths $\mathcal{Y} \in \{96, 192, 336, 720\}$ and past sequence length $\mathcal{X} = 336$.  The best results are in \textbf{bold}, the second best is \uline{underlined}, and the better results between All-at-Once and Average are \colorbox{lightgray}{shaded}.}
	\centering
	\resizebox{\linewidth}{!}{
		\begin{tabular}{cc|c|cc|cc|cc|ccc}
			\cline{2-11}
			&\multicolumn{2}{c|}{Model}& \multicolumn{2}{c|}{Patchformer}& \multicolumn{2}{c|}{Autoformer}& \multicolumn{2}{c|}{Crossformer}& \multicolumn{2}{c}{Transformer}&  \\
			\cline{2-11}
			&\multicolumn{2}{c|}{Metric}&MSE&MAE&MSE&MAE&MSE&MAE&MSE&MAE\\
			\cline{2-11}
			&\multirow{4}*{\rotatebox{90}{All-at-Once}}& 96    & \textbf{0.113} & \textbf{0.255} & \colorbox{lightgray}{0.139} & \colorbox{lightgray}{0.286} & \uline{0.130} & \uline{0.274} & \colorbox{lightgray}{0.146} & \colorbox{lightgray}{0.282}  \\
            &\multicolumn{1}{c|}{}& 192   & \textbf{0.147} & \textbf{0.292} &\colorbox{lightgray}{\uline{0.156}} & \colorbox{lightgray}{\uline{0.304}} & \colorbox{lightgray}{0.173} & \colorbox{lightgray}{0.318} & \colorbox{lightgray}{0.233} & \colorbox{lightgray}{0.357} \\
            &\multicolumn{1}{c|}{}& 336   & \colorbox{lightgray}{\textbf{0.179}} & \colorbox{lightgray}{\textbf{0.324}} & \colorbox{lightgray}{\uline{0.187}} & \uline{0.336} & 0.368 & 0.467 & \colorbox{lightgray}{0.318} & \colorbox{lightgray}{0.408}  \\
            &\multicolumn{1}{c|}{}& 720   & \colorbox{lightgray}{\uline{0.255}} & \colorbox{lightgray}{\uline{0.389}} & \colorbox{lightgray}{\textbf{0.234}} & \colorbox{lightgray}{\textbf{0.378}} & \colorbox{lightgray}{0.499} & \colorbox{lightgray}{0.540} & 0.471 & 0.501  \\
			\cline{2-11}
			&\multirow{4}*{\rotatebox{90}{Electricity}}& 96    & \textbf{0.115} & \textbf{0.263} & 0.154 & 0.306 & \uline{0.134} & \uline{0.284} & 0.186 & 0.321  \\
            &\multicolumn{1}{c|}{} & 192   & \textbf{0.145} & \textbf{0.297} & \uline{0.161} & \uline{0.311} & 0.230 & 0.375 & 0.255 & 0.374  \\
            &\multicolumn{1}{c|}{}& 336   & \textbf{0.178} & \textbf{0.330} & \uline{0.209} & \uline{0.355} & 0.246 & 0.391 & 0.491 & 0.518  \\
            &\multicolumn{1}{c|}{}& 720   & \uline{0.311} & \uline{0.440} & \textbf{0.289} & \textbf{0.427} & 0.584 & 0.599 & 0.539 & 0.554  \\
            \cline{2-11}
			&\multirow{4}*{\rotatebox{90}{Gas}}& 96    & \textbf{0.073} & \textbf{0.198} & 0.111 & 0.253 & \uline{0.088} & \uline{0.216} & 0.101 & 0.233  \\
			&\multicolumn{1}{c|}{}& 192   & \textbf{0.085} & \textbf{0.221} & 0.131 & 0.281 & \uline{0.114} & \uline{0.253} & 0.159 & 0.294  \\
			&\multicolumn{1}{c|}{}& 336   & \textbf{0.101} & \textbf{0.241} & \uline{0.128} & \uline{0.279} & 0.174 & 0.315 & 0.192 & 0.320  \\
			&\multicolumn{1}{c|}{}& 720   & \textbf{0.137} & \textbf{0.287} & \uline{0.196} & \uline{0.330} & 0.382 & 0.467 & 0.210 & 0.338  \\
			\cline{2-11}
			&\multirow{4}*{\rotatebox{90}{GHG}}& 96    & \textbf{0.139} & \textbf{0.288} & 0.171 & 0.323 & \uline{0.154} & \uline{0.305} & 0.188 & 0.333  \\
            &\multicolumn{1}{c|}{}& 192   & \textbf{0.206} & \textbf{0.355} & \uline{0.211} & 0.362 & 0.212 & \uline{0.359} & 0.360 & 0.453  \\
            &\multicolumn{1}{c|}{}& 336   & 0.312 & \uline{0.438} & \textbf{0.225} & \textbf{0.370} & \uline{0.311} & 0.439 & 0.432 & 0.492  \\
            &\multicolumn{1}{c|}{}& 720   & \uline{0.439} & \uline{0.522} & \textbf{0.273} & \textbf{0.411} & 0.682 & 0.643 & 0.551 & 0.565  \\
			\cline{2-11}
            &\multirow{4}*{\rotatebox{90}{Average}}& 96    & \colorbox{lightgray}{\textbf{0.109}} & \colorbox{lightgray}{\textbf{0.250}} & 0.145 & 0.294 & \colorbox{lightgray}{\uline{0.125}} & \colorbox{lightgray}{\uline{0.268}} & 0.159 & 0.295  \\
            &\multicolumn{1}{c|}{} & 192    & \colorbox{lightgray}{\textbf{0.145}} & \colorbox{lightgray}{\textbf{0.291}} & \uline{0.168} & \uline{0.318} & 0.185 & 0.329 & 0.258 & 0.374  \\
            &\multicolumn{1}{c|}{}& 336    & \uline{0.197} & \uline{0.336} & \textbf{0.187} & \colorbox{lightgray}{\textbf{0.335}} & \colorbox{lightgray}{0.244} & \colorbox{lightgray}{0.382} & 0.371 & 0.443  \\
            &\multicolumn{1}{c|}{}& 720    & \uline{0.296} & \uline{0.416} & \textbf{0.252} & \textbf{0.389} & 0.549 & 0.570 & \colorbox{lightgray}{0.433} & \colorbox{lightgray}{0.486}  \\
			\cline{2-11}
		\end{tabular}
  }
	\label{tab:multi-energy forecasting comparison}
\end{table}
\linespread{1}

\subsection{Different Length of Past Time Sequences}\label{sec:diff past sequence}

\begin{figure}
	\includegraphics[width=1\linewidth]{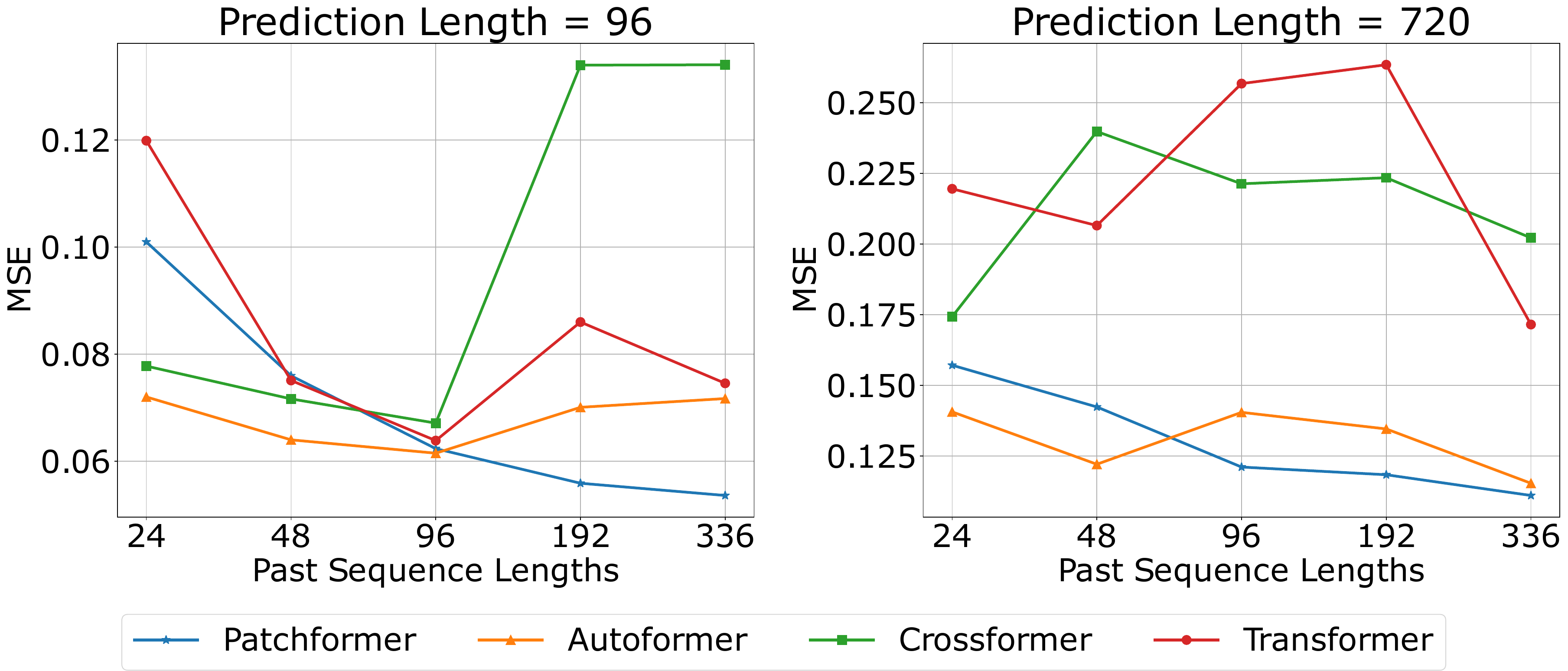}
	\caption{Performance analysis with different past time sequence lengths on Multi-Energy dataset.}
	\label{fig:past_lengths}
\end{figure} 

\begin{table}[t]
\small
\caption{Multi-Energy load forecasting results with Patchformer on Multi-Energy dataset. We use prediction lengths $\mathcal{Y} \in \{96, 720\}$ and past sequence lengths $\mathcal{X} \in \{24, 48, 96, 192, 336\}$.  The best results are in \textbf{bold} and the second best are \uline{underlined}.}
	\centering
	\resizebox{\linewidth}{!}{
		\begin{tabular}{cc|c|cc|cc|cc|ccc}
			\cline{2-11}
			&\multicolumn{2}{c|}{Model}& \multicolumn{2}{c|}{Patchformer}& \multicolumn{2}{c|}{Autoformer}& \multicolumn{2}{c|}{Crossformer}& \multicolumn{2}{c}{Transformer}&  \\
			\cline{2-11}
			&\multicolumn{2}{c|}{Metric}&MSE&MAE&MSE&MAE&MSE&MAE&MSE&MAE\\
			\cline{2-11}
			&\multirow{5}*{\rotatebox{90}{Pred: 96}}& 24    & 0.101 & 0.170 & \textbf{0.072} & \textbf{0.155} & \uline{0.078} & \uline{0.158} & 0.120 & 0.182  \\
            &\multicolumn{1}{c|}{}& 48  & 0.076 & \uline{0.150} & \textbf{0.064} & \textbf{0.147} & \uline{0.072} & 0.155 & 0.075 & 0.151 \\
            &\multicolumn{1}{c|}{}& 96   & \uline{0.062} & \textbf{0.135} & \textbf{0.061} & \uline{0.150} & 0.067 & 0.150 & 0.063 & 0.147  \\
            &\multicolumn{1}{c|}{}& 192  & \textbf{0.056} & \textbf{0.128} & \uline{0.070} & 0.167 & 0.134 & 0.212 & 0.086 & \uline{0.165}  \\
            &\multicolumn{1}{c|}{}& 336   & \textbf{0.054} & \textbf{0.126} & \uline{0.072} & 0.162 & 0.134 & 0.214 & 0.075 & \uline{0.160}  \\
			\cline{2-11}
			&\multirow{5}*{\rotatebox{90}{Pred: 720}}& 24    & \uline{0.158} & \uline{0.224} & \textbf{0.141} & \textbf{0.224} & 0.174 & 0.250 & 0.220 & 0.256  \\
            &\multicolumn{1}{c|}{} & 48   & \uline{0.142} & \uline{0.214} & \textbf{0.122} & \textbf{0.210} & 0.240 & 0.287 & 0.207 & 0.244  \\
            &\multicolumn{1}{c|}{}& 96   & \textbf{0.121} & \textbf{0.193} & \uline{0.140} & \uline{0.215} & 0.221 & 0.274 & 0.257 & 0.273  \\
            &\multicolumn{1}{c|}{}& 192   & \textbf{0.118} & \textbf{0.191} & \uline{0.135} & \uline{0.216} & 0.223 & 0.282 & 0.263 & 0.280  \\
            &\multicolumn{1}{c|}{}& 336   & \textbf{0.111} & \textbf{0.183} & \uline{0.115} & \uline{0.223} & 0.202 & 0.289 & 0.172 & 0.226  \\
            \cline{2-11}
		\end{tabular}
  }
	\label{tab:Past length vary}
\end{table}

In this section, the forecasting performance between Patchformer and other models with different lengths of past time sequences has been compared and shown in Figure \ref{fig:past_lengths} and Table \ref{tab:Past length vary}. Five different past sequence lengths are selected: $\{24, 48, 96, 192, 336\}$. Two prediction lengths are $\{ 96, 720 \}$. Intuitively, the models' forecasting performance and the length of past sequences should be positively correlated. However, based on the argument in \cite{zeng2023transformers}, this principle may not work for the majority of the Transformer-based models since they cannot capture the temporal local information efficiently. Figure \ref{fig:past_lengths} also proves the phenomenon, which shows that except for the proposed Patchformer, all other models do not follow the positive correlation between model performance and the past sequence length. This shows Patchformer's ability to capture long-range, past local semantic information. In addition, Table \ref{tab:Past length vary} shows that, as the length of the past sequence increases, Patchformer's performance improves and surpasses all other models.  

\section{Conclusion} \label{sec:Conclusion}
The paper presents Patchformer, a novel Transformer-based model for long-term multi-energy load forecasting. It addresses the challenges in IMES forecasting by predicting each feature in multivariate time-series data independently and segmenting it into patches, which can benefit the capture of interdependence among different features and the reception of local semantic information. The Patchformer's architecture, including its patch embedding and encoder-decoder mechanism, is illustrated in detail. In numerical analysis, the Patchformer model demonstrates superior performance against other state-of-the-art models in multivariate long-term forecasting by the unique Multi-Energy dataset and six other benchmark datasets. Furthermore, the model's performance in univariate long-term forecasting when predicting electricity and gas load is superior to other models. The experiment also demonstrates the positive effect of the interdependence among energy-related products on the performance of the time-series forecasting in the Multi-Energy dataset across Patchformer and other models by comparing the forecasting results between predicting the electricity, gas load and GHG emissions all at once and the average of the individual predictions. In addition, the positive correlation between Patchformer's performance and the past sequence length is stated, which shows its capability to capture long-range past local semantic information. Future research directions can be addressed as follows: handling volatile data, like in financial markets, capturing long-term dependencies, and adapting to the non-stationarity of real-world datasets, which need to be improved in further study. In addition, to further measure the Patchformer performance and allow users to better understand the model's behaviour, interpret and trust the results and output, Explainable AI (XAI) methods, such as SHapley Additive exPlanations (SHAP) \cite{lundberg2017unified} and local interpretable model-agnostic explanations (LIME) \cite{ribeiro2016should} could be implemented for future study.     

\printbibliography[heading=bibintoc]

\end{document}